\documentclass{article}

\PassOptionsToPackage{sort, numbers}{natbib}

\usepackage[final]{nips_2016}

\usepackage[utf8]{inputenc}
\usepackage[T1]{fontenc}
\usepackage{hyperref}
\usepackage{url}
\usepackage{booktabs}
\usepackage{amsfonts}
\usepackage{nicefrac}
\usepackage{microtype}
\usepackage{grffile}
\usepackage{amsmath}
\usepackage{graphicx}
\usepackage{verbatim}
\usepackage{setspace}
\usepackage{subcaption}

\title{Spatio-Temporal Image Boundary Extrapolation}

\author{
  Apratim Bhattacharyya \\
  Max Planck Institute for Informatics \\
  Saarbr\"{u}cken, Germany \\
  \texttt{abhattac@mpi-inf.mpg.de} \\
  \And
  Mateusz Malinowski \\
  Max Planck Institute for Informatics \\
  Saarbr\"{u}cken, Germany \\
  \texttt{mmalinow@mpi-inf.mpg.de} \\
  \AND
  Mario Fritz \\
  Max Planck Institute for Informatics \\
  Saarbr\"{u}cken, Germany \\
  \texttt{mfritz@mpi-inf.mpg.de} \\
}

\begin{document}

\maketitle

\begin{abstract}
Boundary prediction in images as well as video has been a very active topic of research and organizing visual information into boundaries and segments is believed to be a corner stone of visual perception. While prior work has focused on predicting boundaries for observed frames, our work aims at predicting boundaries of future unobserved frames. This requires our model to learn about the fate of boundaries and extrapolate motion patterns. We experiment on established real-world video segmentation dataset, which provides a testbed for this new task. We show for the first time spatio-temporal boundary extrapolation in this challenging scenario. Furthermore, we show long-term prediction of boundaries in situations where the motion is governed by the laws of physics. We successfully predict boundaries in a billiard scenario without any assumptions of a strong parametric model or any object notion. We argue that our model has with minimalistic model assumptions derived a notion of ``intuitive physics'' that can be applied to novel scenes.
\end{abstract}

\section{Introduction}

Humans possess the skill to imagine future states of on observed scene. Part of this skill is footed on an intuitive understanding of physics \cite{Smith1994,mccloskey1983intuitive}. Observing a moving ball,  we have a reasonably good estimate about the future trajectory of the ball, which lets e.g. a goal keeper catch the ball. Similarly, on a game of billiard we have to choose our action under the prediction of future states of the table in order to win the game.  These skills are key to interacting with such dynamic objects that have deterministic fate and let humans excel at such complicated tasks.

Most interestingly, it has been argued that we perform this type of predictions without necessarily using an explicit or formal understanding of the physics which gives rise to these phenomenon. Humans seem to acquire this particular type of physical understanding in a data-driven, learning-based approach from prior experiences \cite{baillargeon1994infants,baillargeon1995model,baillargeon2002acquisition}.

Recently, there has been an increased interest in modeling and predicting phenomena that are governed by the laws of physics \cite{wu2015galileo,li2016fall,mottaghi2015newtonian,fragkiadaki2015learning}. These models typical are parametric of some sort or only predict a qualitative outcome of the scene.
More generally, full future frame predication has been studied that is agnostic to the underlying cause of the change depicted in the sequence \cite{ranzato2014video}. In contrast to the physical models, only very short range predictions have been shown and there are blurriness problems with predicting the full future appearance. \cite{lerer2016learning} has predicted future segmentation masks, but also suffers from blurriness in the prediction.

Recently a lot of progress has been made in the field of video segmentation supported by datasets like VSB100 \citep{galasso2013unified}. The performance of any video segmentation algorithm is measured with respect to a set of human annotations. Humans tend to annotate semantically coherent objects and regions as segments. The segmented natural videos discard many details of natural videos mentioned above which are hard for a model to learn and still captures the important objects as boundaries. The boundaries between segments gives rise to boundary images.

Our main contribution is the first model to predict future boundary frames of segmented videos and to explore the performance of these models to under structured motion -- physical and non-physical ones. We evaluate performance both on real world videos from VSB100 and synthetic videos. Moreover, we show that our models can develop an intuitive understanding of physics from raw visual input without any strong parametric model of the motion or ``object notion". That is, the model does not know a priori the location or type of the objects it is supposed to model.

\begin{figure}[tb]
    \centering
    \includegraphics[width = \textwidth]{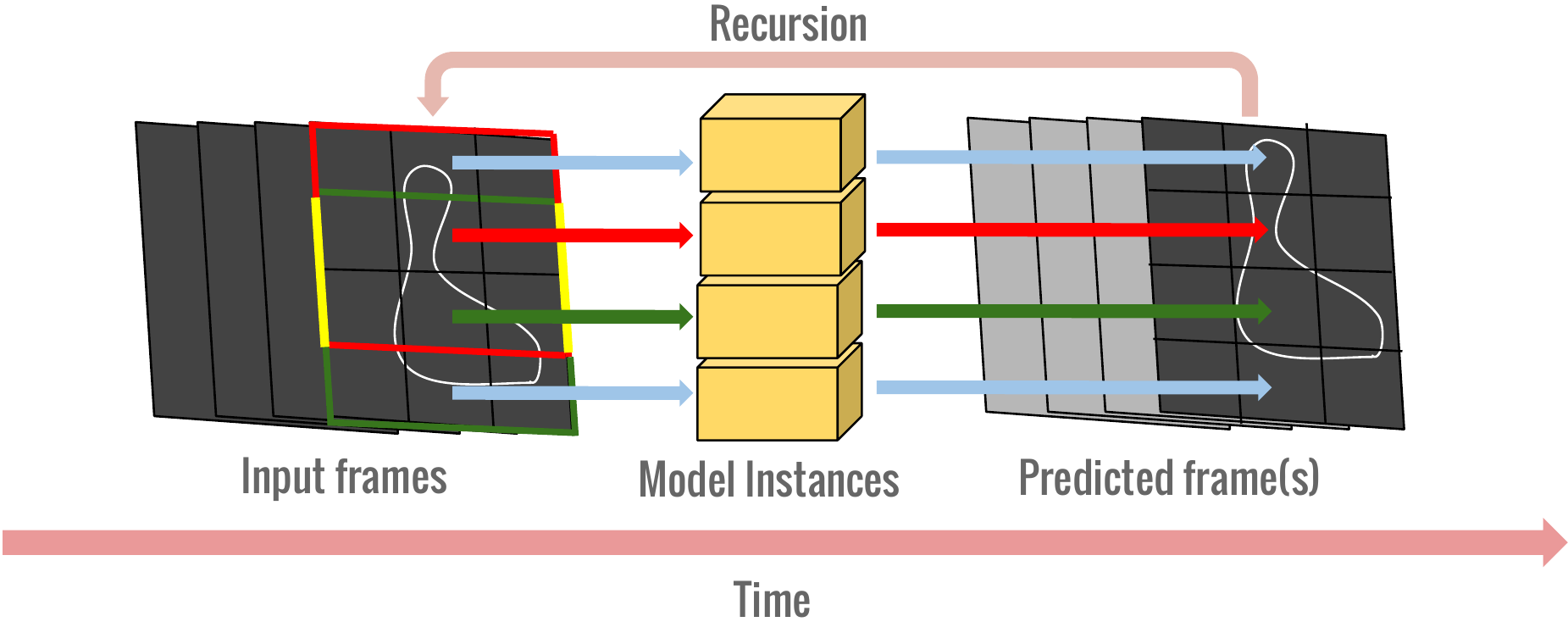}
    \caption{General framework that encapsulates all our architectures.}
    \label{fig:framework}
\end{figure}

\section{Related work}
{\bf Frame prediction.}
This problem has been recently explored in \citep{srivastava2015unsupervised, ranzato2014video, mathieu2015deep}. The work of \citet{srivastava2015unsupervised} focused on learning representations of video sequences. They used an LSTM encoder unit to encode videos into a vector which they used for predicting future frames. In \citet{ranzato2014video} the authors focused on the problem of blurring caused by using the mean squared loss as an objective function. They sought to remedy this problem by discretizing the input through k-means atoms and predicting on this vocabulary instead. The work of \citet{mathieu2015deep} also focused on this problem. They proposed using adversarial loss, which lead to improved results over \citet{ranzato2014video}. These works have focused on natural videos or datasets like MNIST digits. Also, the ability of the models developed in the previously mentioned works to learn the dynamics of structured motion has not been explored. Works of \citet{sutskever2009recurrent, michalski2014modeling} generate frames of videos of bouncing balls, but their dataset is very limited in size and resolution. Moreover, they do not consider generalization. \\
{\bf Equation parameter estimation.}
Works like that of \citet{wu2015galileo, mottaghi2015newtonian} focus on predicting outcomes of physical events in videos or images. In \citep{wu2015galileo} the authors propose ``Galileo'' that estimates the physical properties of objects and inverts a physics engine to predict outcomes. While in \citep{mottaghi2015newtonian}, the authors predict the motion of objects using a single query image using a neural network which matches the image to a moment in a video which is closest in describing the dynamics of motion of the scene depicted in the image.\\
{\bf Intuitive physics.}
The ability of artificial neural networks to develop such an intuitive understanding of physics from raw visual input has been recently explored in \citep{fragkiadaki2015learning, lerer2016learning, li2016fall}. \citet{fragkiadaki2015learning} had developed a model which could predict futures states of balls moving on a billiard table. Whereas, \citet{lerer2016learning} and \citet{li2016fall} had developed a model which could predict the stability of towers made out of blocks. The work of \citet{lerer2016learning} could also predict future locations of the blocks. However, both \citet{fragkiadaki2015learning} and \citet{lerer2016learning} have an ``object notion''. \citet{li2016fall} focuses only on predicting the outcome not the exact state. \\
{\bf Video segmentation. }
Video segmentation as the task of finding consistent spatio-temporal boundaries in a video volume has received significant attention over the last years \citep{galasso2014spectral,ochs2014segmentation,galasso2013unified,chang2013video}, as it provides an initial analysis and abstraction for further processing. In contrast, our approach aims at extrapolating these boundaries into the future without any video observed for the future frames. Our proposed model could serve as an expectation on boundaries in future frames and therefore help improve video segmentation prediction in future work.

\section{Models}

In order to extrapolate spatio-temporal boundaries to future unobserved frames, we are building on the recent success of deep learning that allows to construct flexible and end-to-end trainable architectures with strong visual feature encoders, predictors and the recent success of formulating decoders which reconstruct a target image. We are facing several key challenges that will be reflected in the design choices of our architecture:

\noindent
\label{ssec:designchoices}
{\bf Large Spatio-Temporal Receptive Field. } The output layer neurons should have a wide receptive field to preserve long range spatial and temporal dependencies and learn about interaction with other boundaries in a spatio-temporal context. Therefore we explore multiple convolutional layers optionally with pooling or with fully connected layers.\\
{\bf Preserving Resolution and Preventing Blurred Output. } The models must maintain resolution in order to derive a high fidelity output boundary map. Excessive pooling or tight bottlenecks with fully connected layers have been shown successful in classification tasks, but also have shown to induce image degradations for image synthesis tasks \citep{ranzato2014video}.\\
{\bf Varying resolution. } In order to facilitate extrapolation in different scenarios with changing resolutions and aspect ratios, we target a patch based approach. Such an approach has been successfully used by \citep{ranzato2014video, mathieu2015deep}. Moreover, such an approach simplifies the learning problem by reducing the input dimensionality and modelling reoccurring motion patterns locally. \\ 
{\bf Extrapolation over long time scales. } In order to make local predication globally consistent, we have to introduce spatio-temporal dependencies. Consider a video of a moving ball. The trajectory of a ball might intersect with multiple patches. To correctly extrapolate the motion far into the future, instances of a model predicting on neighboring patches need to communicate. In order to allow for exchange of information between different patch extrapolators, we will experiment with larger overlapping receptive fields, that constitute a read-write architecture, where the past frames (observed or already extrapolated ones) serve as a kind of shared memory. However, this potentially makes the learning problem more complex.

\subsection{Prediction modes}
\label{ssec:Prediction Modes}
Here we consider two modes for predicting future frames:\\
{\bf Sequence to sequence. } Here a model takes a sequence of past frames and predicts another sequence. This mode of prediction has been widely explored in machine translation \citep{cho2014learning} where the input and output sequences are natural language sentences. \citet{srivastava2015unsupervised} applied this framework to sequence of images. However, output sequence length is limited by memory and processing time and communication between the patches during extrapolation becomes impossible.\\
{\bf Recursive. } A model which predicts only one future frame can be made to predict sequences by sampling from its output and using the output as input in the next time step. This mode allows the instances of the model predicting on neighbouring patches to communicate (neighbouring patches are also given as input to model instance i.e. a context). However, in \citet{mathieu2015deep} this mode led to better short term but worse long term results. This is understandable as the model does not get to learn the long term dynamics of motion.

\subsection{Model architectures}
\label{ssec: model_arch}
We propose the following architectures for boundary extrapolation while keeping in mind the key challenges outlined previously. They all fit in the general framework outlined in \autoref{fig:framework} and can operate in both the prediction modes outlined previously.

\begin{figure}
    \centering
    \begin{subfigure}{\textwidth}
        \centering
        \includegraphics[width=1.0\textwidth]{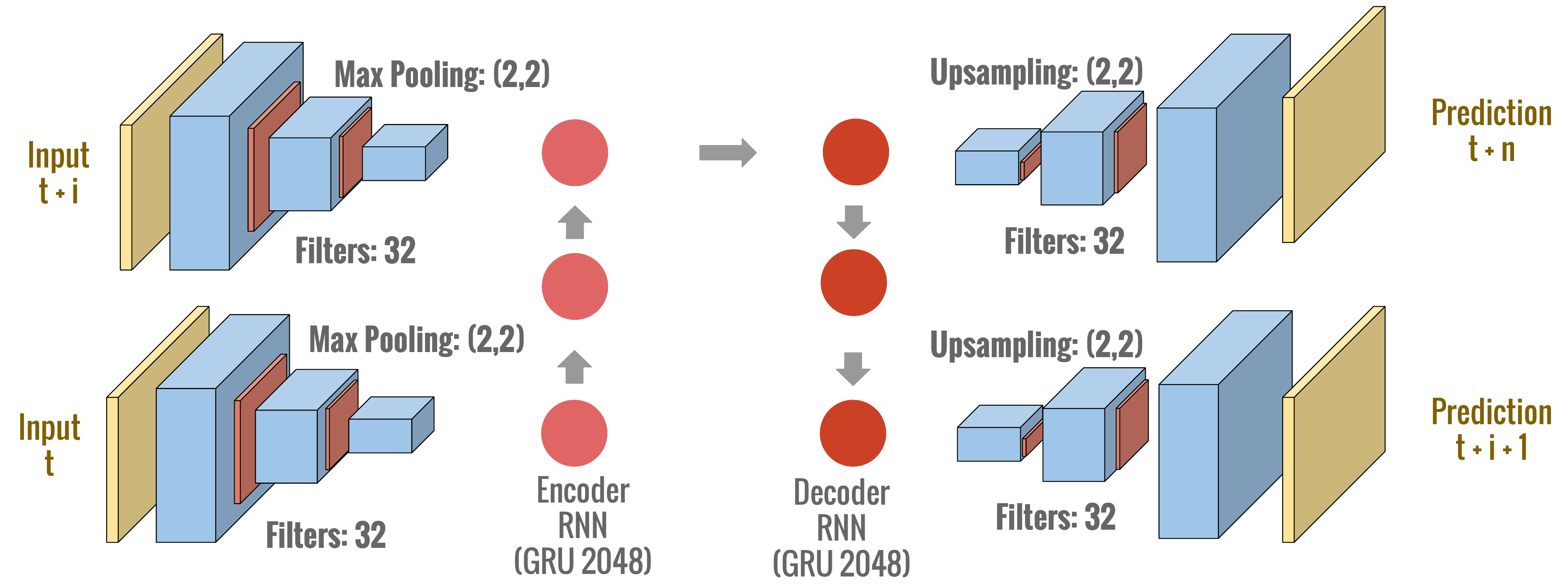}    
        \caption{Convolutional RNN - encoder decoder architecture}
        \label{fig:tdconv}
    \end{subfigure}
    \begin{subfigure}{\textwidth}
        \centering
        \includegraphics[width=1.0\textwidth]{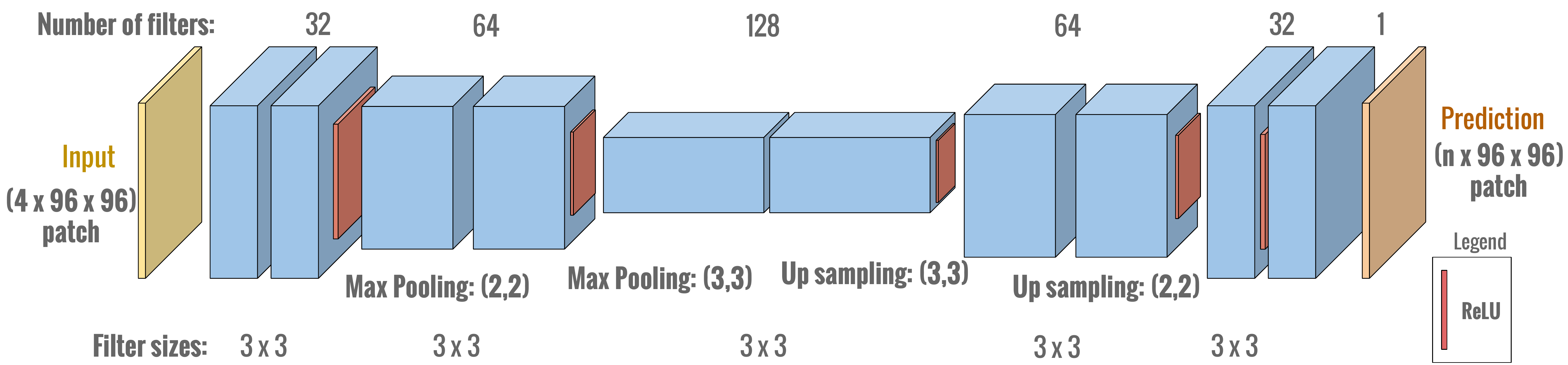}    
        \caption{Fully-Convolutional architecture}
        \label{fig:conv}
    \end{subfigure}
    \begin{subfigure}{\textwidth}
        \centering
        \includegraphics[width=1.0\textwidth]{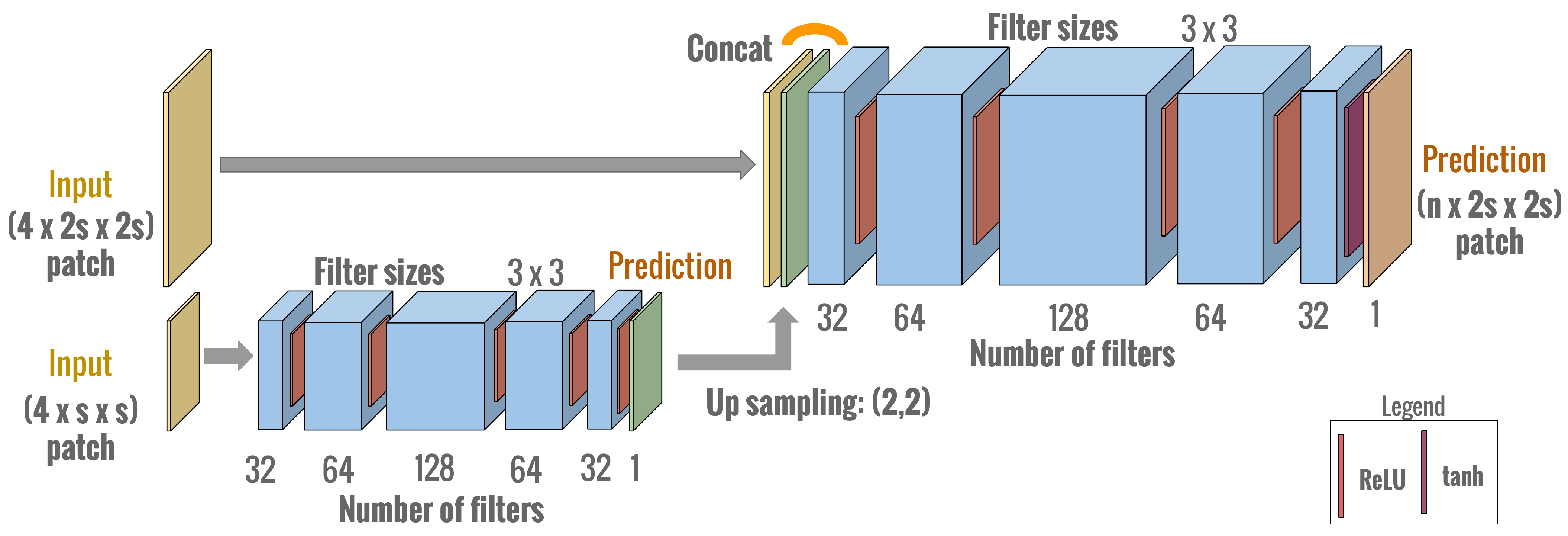}    
        \caption{Multi-scale architecture}
        \label{fig:ms}
    \end{subfigure}
    \caption{Our concrete architectures.}
    \label{fig:modelarch}
\end{figure}

\noindent
{\bf RNN - encoder decoder. } This model architecture has been explored in \citet{srivastava2015unsupervised}. This architecture consists of an encoder LSTM unit which reads in the input frame sequence one time step at a time and produces a single vector as output. This vector is then read by either a dense layer with a non-linearity to produce the final output in the recursive mode or by a decoder LSTM to produce the output sequence in the sequence to sequence mode. Here, we have used GRU units \citep{chung2014empirical} instead of LSTM units because they provide faster convergence without loss of performance.

{\bf Convolutional RNN - encoder decoder.}
Convolutional Neural Networks with convolutional layers which extract high quality location invariant features have been very successful in various tasks such as object recognition \citep{jarrett2009best}. We extend the architecture described previously with convolutional layers to extract features along with ReLU non-linearities \citep{nair2010rectified} and pooling layers to extract features which is then fed into a encoder GRU unit (see \autoref{fig:tdconv}). To ensure output resolution we up-sample the output of the GRU unit followed by convolution. This can be thought of as convolution with a fractional stride.

{\bf Fully Convolutional. } Both the previously discussed architectures have output neurons with wide receptive fields but they are autoencoders \citep{srivastava2015unsupervised} which involve compression. One way to deal with this problem is to increase the size of the encoder (decoder). Another way is to use end to end convolutions instead. For the output neurons of such a model to have a large perceptive field will require many layers of convolutions, making it computationally expensive. We can try to balance both these factors by using moderate number of pooling layers and an increasing number of filters (with ReLU) and upsampling followed by convolution to maintain resolution (see \autoref{fig:conv}).

{\bf Multi-Scale. }
Another approach to maintain resolution and preserve long range spatial dependencies is to use a multi-scale architecture akin to a Laplacian pyramid \citep{denton2015deep}. Such an architecture has been used successfully for generating natural images \citep{denton2015deep} and predicting future natural frames \citep{mathieu2015deep}.  We use four scales, each downsampled by a factor of two from the one before. Each each level captures image structure present at a particular scale. This creates a wide receptive field for the output layer neurons without pooling.  We use a five layer convolutional network at each scale with ReLU non-linearities and a tanh non-linearity at the end. The output at a certain scale is upsampled and used as input at the next larger scale as a candidate future frame (see \autoref{fig:ms}).

\subsection{Loss function}
A variety of loss functions have been tried for frame prediction \citep{srivastava2015unsupervised, ranzato2014video, mathieu2015deep}. These include L1, L2 (also MSE or Mean Squared Error), Adversarial, sharpness measures like GDL(Gradient Difference Loss) etc. It has been observed that L2 loss leads to blurry predictions on grayscale or RGB data because in case of natural images it is possible to obtain low error by blurring the input. Adversarial loss produces sharper results which look more similar to natural images.  However, as we want to predict boundary images this effect should be limited. Thus, as a starting point we use the mean squared error.

\section{Experiments}
We evaluate the models in \autoref{ssec: model_arch} first on real data and then on long range extrapolation on sequences with structured, deterministic motion. We convert each video into 32x32 pixel patches. The network observes a patch along with its eight neighbouring patches (when trained with  context) at the current time-step and three time-steps into the past. When we use a context, we let the networks predict the next 96x96 patch(s) and use the middle 32x32 patch as high confidence output and discard the rest. This allows for fair comparison with the multi-scale architecture which has same output resolution as input. We use boundary precision recall (BPR) \citep{galasso2013unified} as the evaluation metric. This metric can be defined for a set $P$ of predicted boundary images and $G$ of corresponding ground truth boundary images as:

\begin{align*}
    P = \frac{ \sum_{B_{p} \in P, B_{g} \in G} \mid B_{p} \cap B_{g} \mid }{\sum_{B_{p} \in P} \mid B_{p} \mid} \;\;\;\;
    R = \frac{ \sum_{B_{p} \in P, B_{g} \in G} \mid B_{p} \cap B_{g} \mid }{\sum_{B_{g} \in G} \mid B_{g} \mid} \;\;\;\;
    F &= \frac{2 PR}{P + R}
\end{align*}

where $P$ is boundary precision, $R$ is boundary recall and $F$ is the combined F-measure.

\subsection{Real data}
\paragraph{Training.}
We use the VSB100 dataset which contains 100 videos with a of maximum 121 frames each.  We randomly choose 30 videos for training and 30 for testing videos from the VSB100 dataset. The videos contains a wide range of objects of different sizes and shapes. The videos also have a wide variety of both object and camera motion. We use the hierarchical video segmentation algorithm in \citet{khoreva2016improved} to segment these videos. The output is a ultra-metric contour map (ucm). Boundaries higher in the hierarchy represent more semantically coherent entities like animals, vehicles etc are therefore more stable temporally. We discard boundaries belonging to the lowest level of the hierarchy as they are very unstable temporally. However, we keep the rest intact and use their values as a confidence measure.

\begin{table}[h]
  \caption{AUC and best F measure on VSB100}
  \label{auc_fm_rec}
  \centering
  \begin{tabular}{lccccc}
    \toprule
    \textbf{Time-steps} &\textbf{Last Input} &\textbf{RNN}   &\textbf{Conv-RNN}  &\textbf{Fully-Conv}  &\textbf{Multi-Scale}  \\
    \midrule
    \multicolumn{6}{c}{AUC: Sequence to Sequence}                   \\
    \midrule
    t + 1      &0.163 &0.101 &0.153 &0.269 &\textbf{0.297} \\
    t + 2      &0.096 &0.072 &0.110 &0.193 &\textbf{0.200} \\
    t + 4      &0.060 &0.039 &0.066 &\textbf{0.116} &\textbf{0.115} \\
    \midrule
    \multicolumn{6}{c}{Best F-measure: Sequence to Sequence}                   \\
    \midrule
    t + 1      &0.398 &0.278 &0.323 &0.423 &\textbf{0.435}  \\
    t + 2      &0.304 &0.240 &0.286 &0.358 &\textbf{0.361}  \\
    t + 4      &0.242 &0.182 &0.233 &\textbf{0.284} &\textbf{0.283} \\
    \midrule
    \multicolumn{6}{c}{AUC: Recursive (Without context)}                   \\
    \midrule
    t + 1      &0.163 &0.135 &0.209 &0.292 &\textbf{0.304} \\
    t + 2      &0.096 &0.068 &0.099 &0.170 &\textbf{0.184} \\
    t + 4      &0.060 &0.027 &0.031 &0.084 &\textbf{0.087} \\
    \midrule
    \multicolumn{6}{c}{Best F-measure: Recursive (Without context)}                   \\
    \midrule
    t + 1      &0.398 &0.314 &0.367 &0.432 &\textbf{0.438} \\
    t + 2      &0.304 &0.229 &0.279 &0.351 &\textbf{0.356} \\
    t + 4      &0.242 &0.141 &0.159 &0.255 &\textbf{0.257} \\
    \midrule
    \multicolumn{6}{c}{AUC: Recursive (With context)}                   \\
    \midrule
    t + 1      &0.163 &0.125 &0.168 &0.286 &\textbf{0.309} \\
    t + 2      &0.096 &0.064 &0.067 &0.159 &\textbf{0.185} \\
    t + 4      &0.060 &0.023 &0.017 &0.062 &\textbf{0.087} \\
    \midrule
    \multicolumn{6}{c}{Best F-measure: Recursive (With context)}                   \\
    \midrule
    t + 1      &0.398 &0.304 &0.341 &0.426 &\textbf{0.439} \\
    t + 2      &0.304 &0.232 &0.229 &0.341 &\textbf{0.357} \\
    t + 4      &0.242 &0.134 &0.097 &0.219 &\textbf{0.258} \\
    \bottomrule
  \end{tabular}
\end{table}

\paragraph{Evaluation.}
As the predicted boundaries have different confidence values, we threshold the predictions before comparison to the ground-truth. We vary the threshold to obtain a precision-recall curve and report the area under the curve (AUC) along with the best F-measure across all thresholds. We evaluate each model under three prediction modes: \textbf{sequence to sequence, recursive (with context) and recursive (without context)}. We always include the last input frame as a baseline. As many boundaries do not change between frames in the videos of VSB100, this is not a bad baseline especially when we are predicting one step into the future. 

\begin{figure}
    \centering
    \begin{subfigure}{0.32\textwidth}
        \centering
        \includegraphics[width=1.0\textwidth]{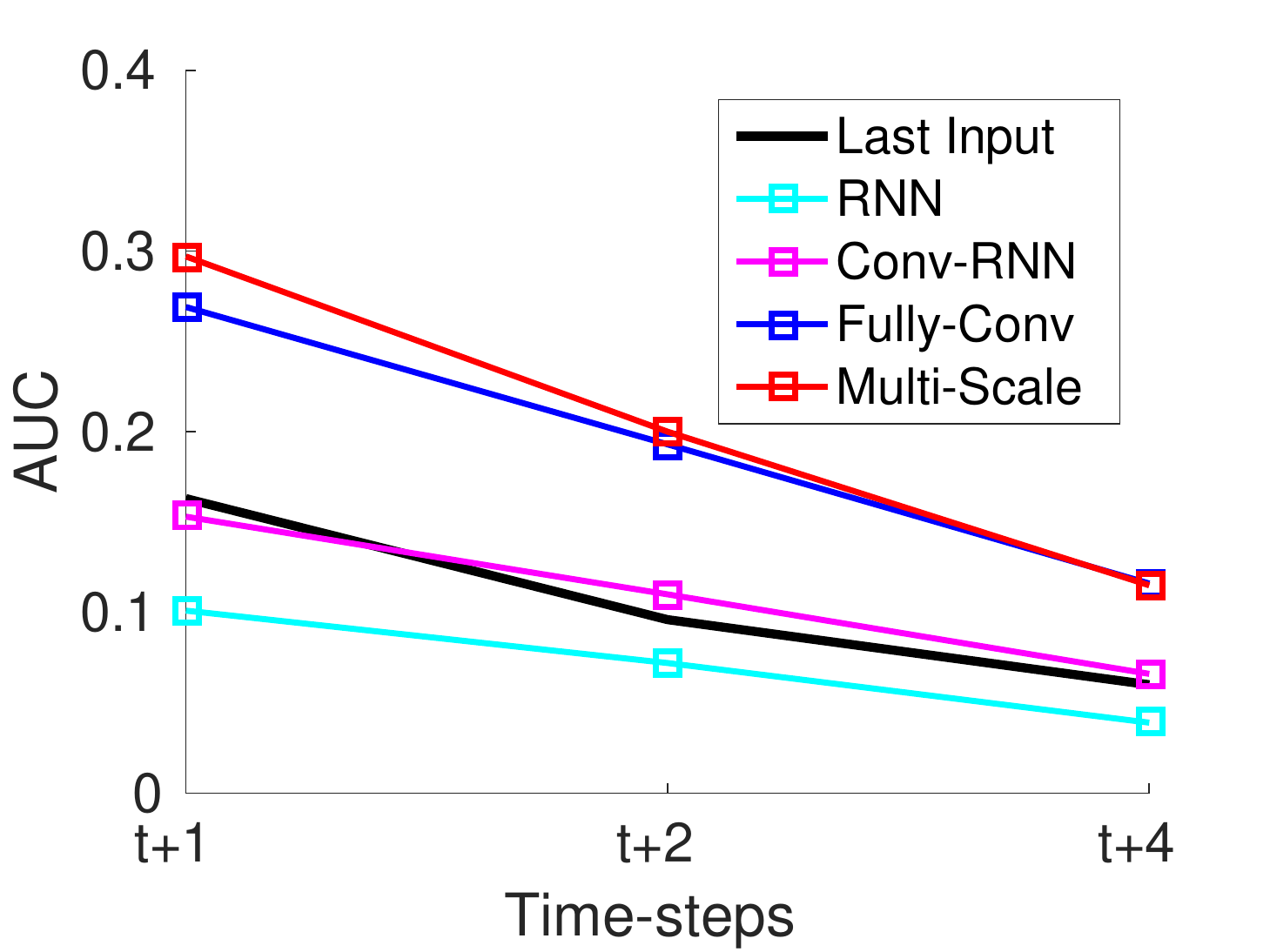}    
    \end{subfigure}
    \begin{subfigure}{0.32\textwidth}
        \centering
        \includegraphics[width=1.0\textwidth]{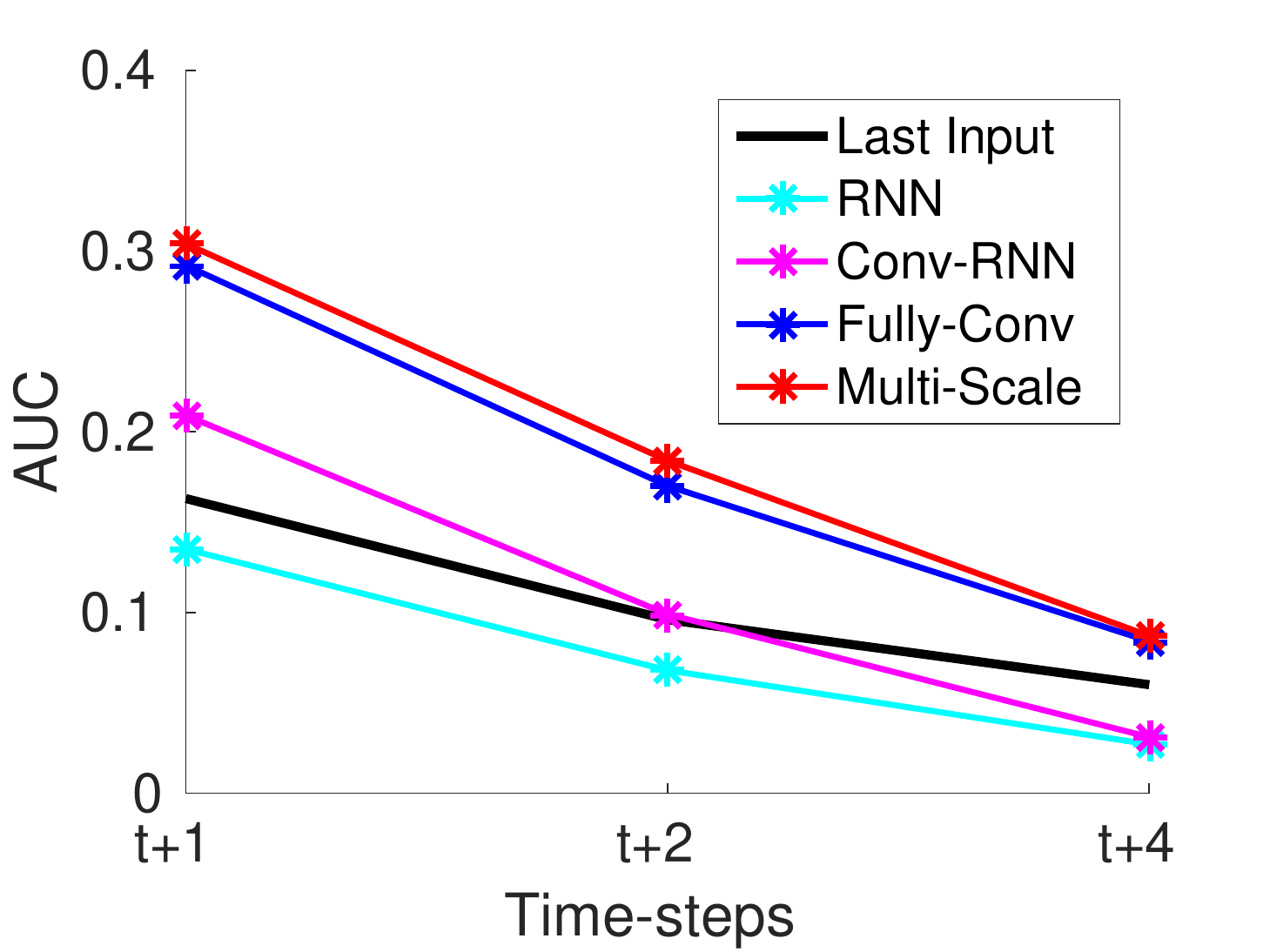}    
    \end{subfigure}
    \begin{subfigure}{0.32\textwidth}
        \centering
        \includegraphics[width=1.0\textwidth]{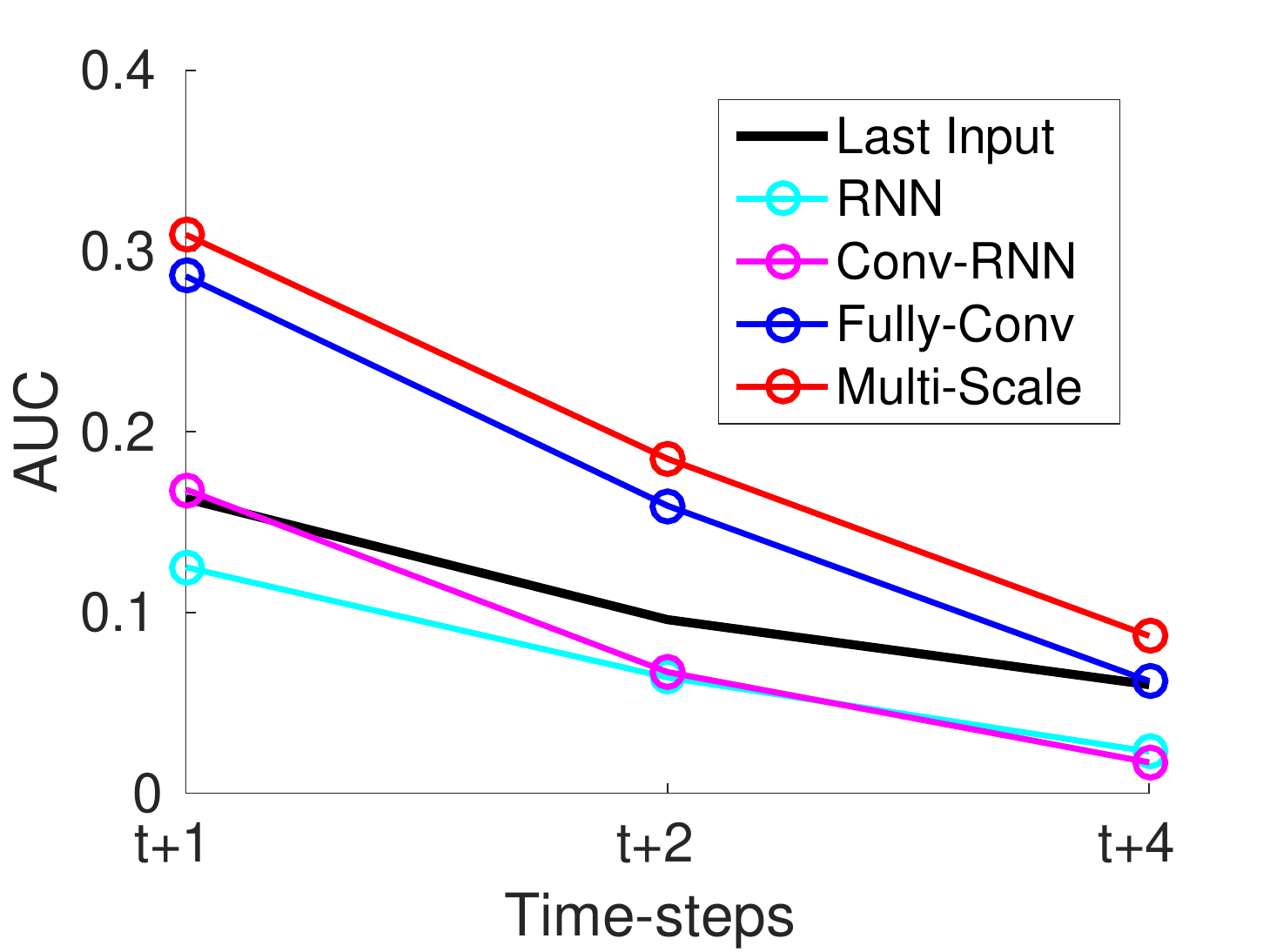}    
    \end{subfigure}
    \label{fig:aucvsb100}
    \begin{subfigure}{0.32\textwidth}
        \centering
        \includegraphics[width=1.0\textwidth]{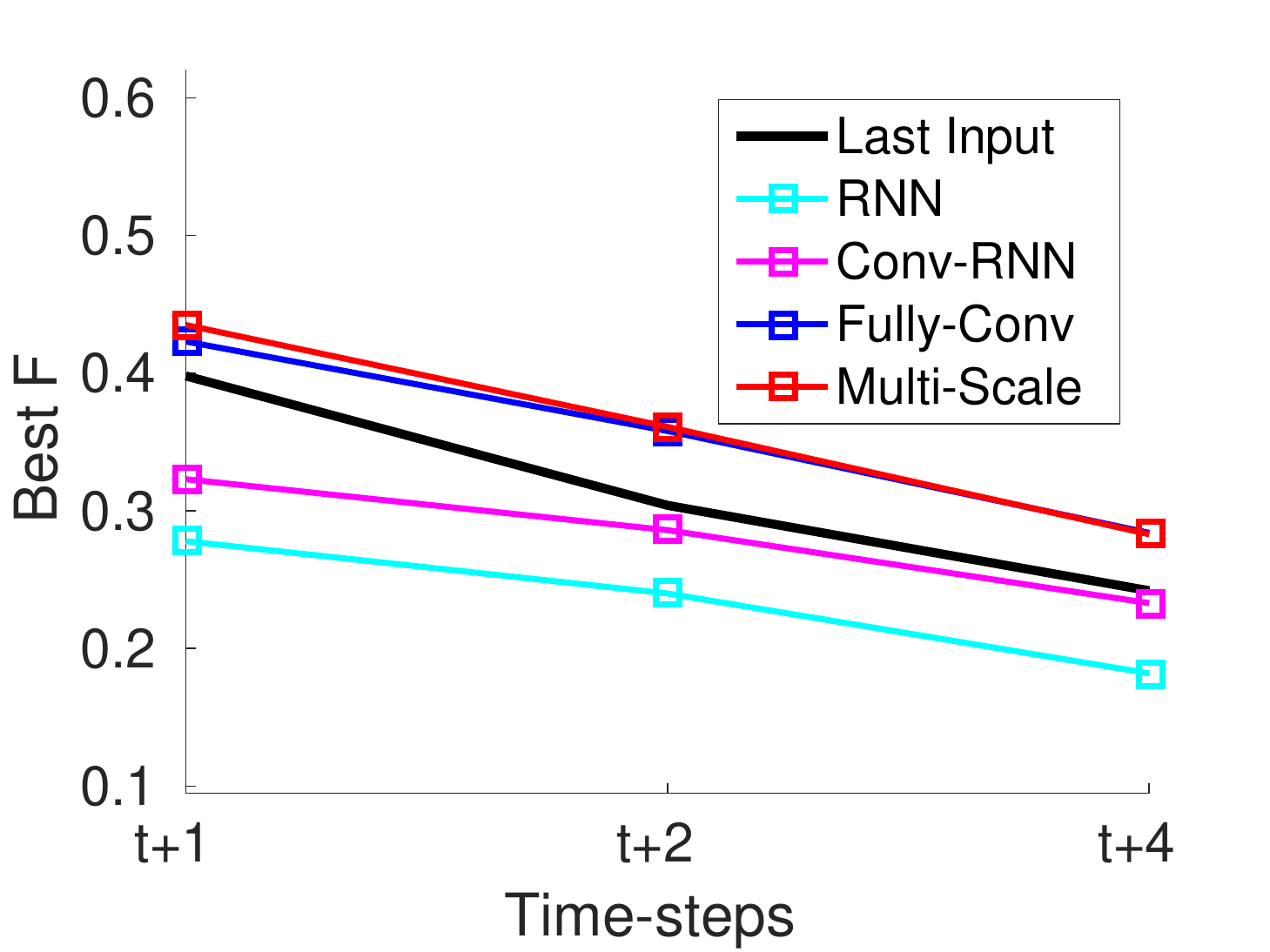}    
    \end{subfigure}
    \begin{subfigure}{0.32\textwidth}
        \centering
        \includegraphics[width=1.0\textwidth]{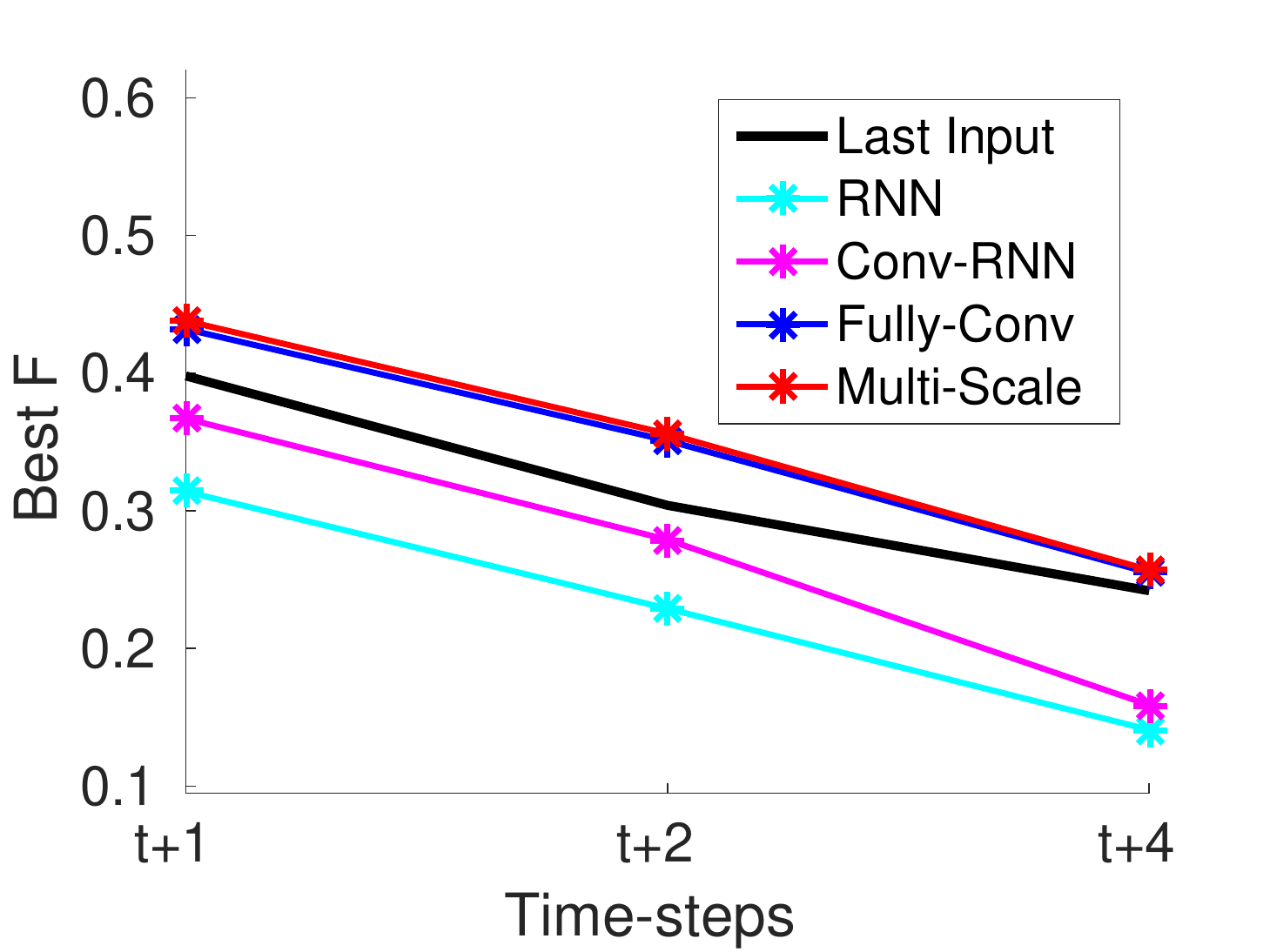}    
    \end{subfigure}
    \begin{subfigure}{0.32\textwidth}
        \centering
        \includegraphics[width=1.0\textwidth]{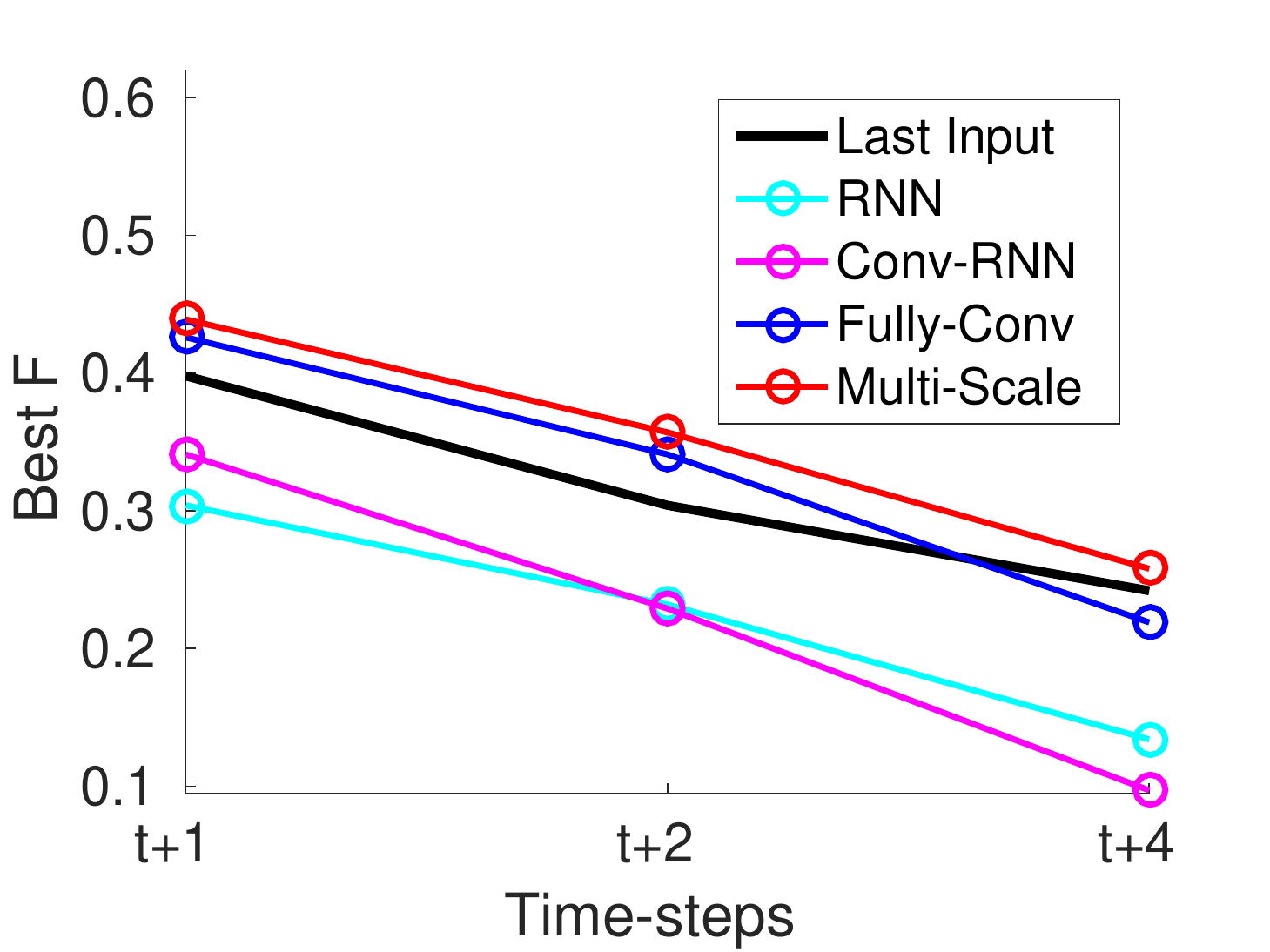}    
    \end{subfigure}
    \caption{Top left to right: AUC for prediction modes: Sequence to Sequence, Recursive (without context), Recursive (with context). Bottom Left to right: Best F-measure for prediction modes: Sequence to Sequence, Recursive (without context), Recursive (with context).}
    \label{fig:aucbestfvsb100}
\end{figure}

\begin{figure}
    \centering
    \begin{minipage}{0.24\textwidth}
        \centering
        \includegraphics[width=1.0\textwidth]{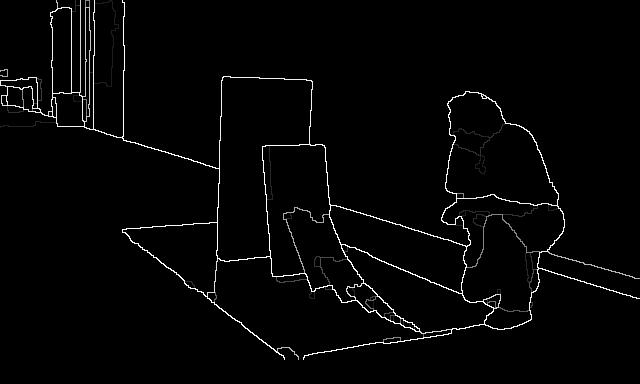}    
        \captionsetup{labelformat=empty,skip=0pt }
        \caption*{Last Input Frame}
    \end{minipage}
    \begin{minipage}{0.24\textwidth}
        \centering
        \includegraphics[width=1.0\textwidth]{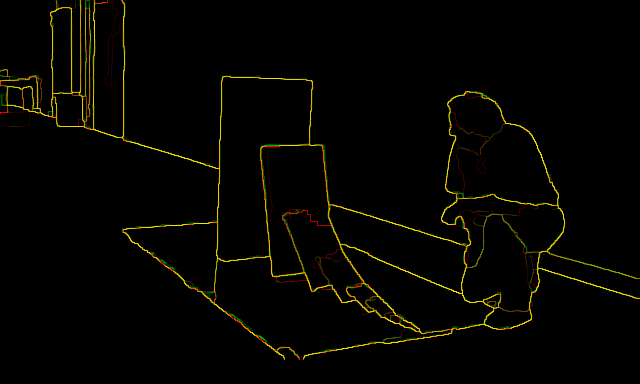}
        \captionsetup{labelformat=empty,skip=0pt }
        \caption*{t + 1}
    \end{minipage}
    \begin{minipage}{0.24\textwidth}
        \centering
        \includegraphics[width=1.0\textwidth]{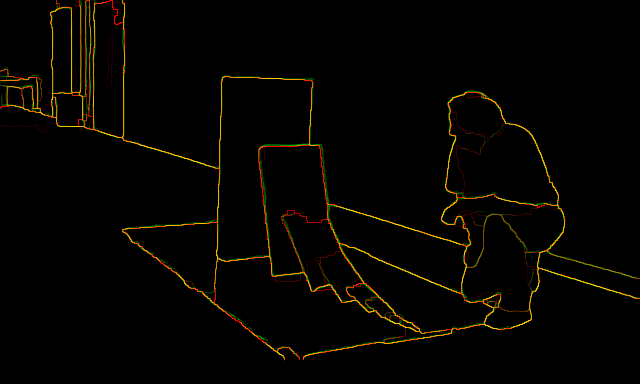}    
        \captionsetup{labelformat=empty,skip=0pt }
        \caption*{t + 2}
    \end{minipage}
    \begin{minipage}{0.24\textwidth}
        \centering
        \includegraphics[width=1.0\textwidth]{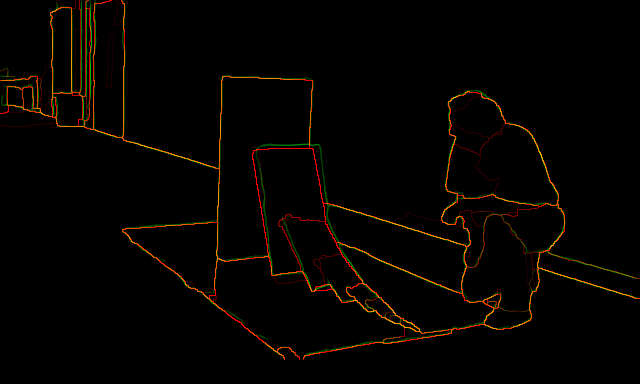}  
        \captionsetup{labelformat=empty,skip=0pt }
        \caption*{t + 4}
    \end{minipage}
    \centering
    \begin{minipage}{0.24\textwidth}
        \centering
        \includegraphics[width=1.0\textwidth]{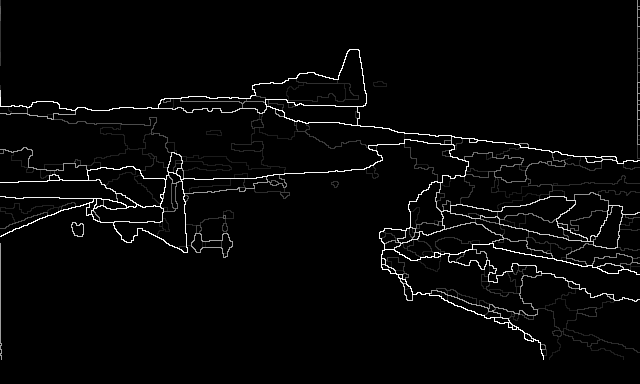}  
        \captionsetup{labelformat=empty,skip=0pt }
        \caption*{Last Input Frame}
    \end{minipage}
    \begin{minipage}{0.24\textwidth}
        \centering
        \includegraphics[width=1.0\textwidth]{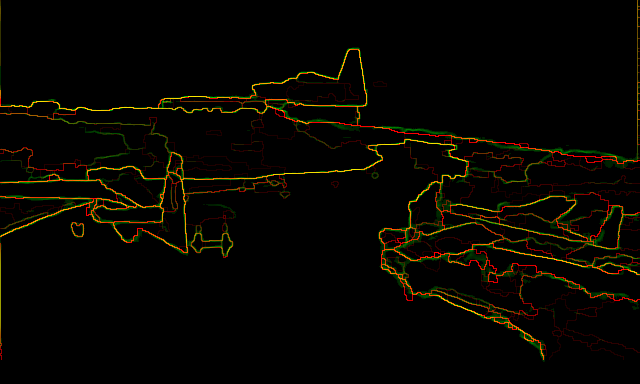}  
        \captionsetup{labelformat=empty,skip=0pt }
        \caption*{t + 1}
    \end{minipage}
    \begin{minipage}{0.24\textwidth}
        \centering
        \includegraphics[width=1.0\textwidth]{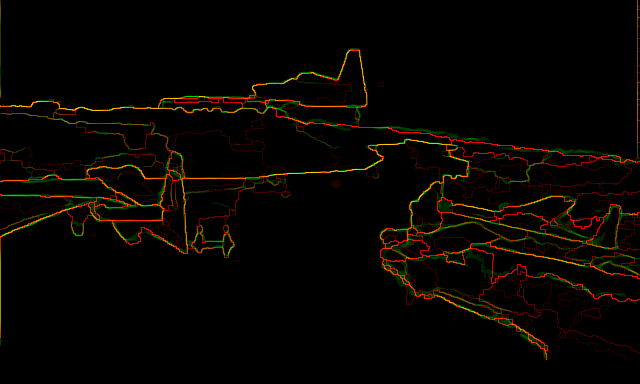} 
        \captionsetup{labelformat=empty,skip=0pt }
        \caption*{t + 2}
    \end{minipage}
    \begin{minipage}{0.24\textwidth}
        \centering
        \includegraphics[width=1.0\textwidth]{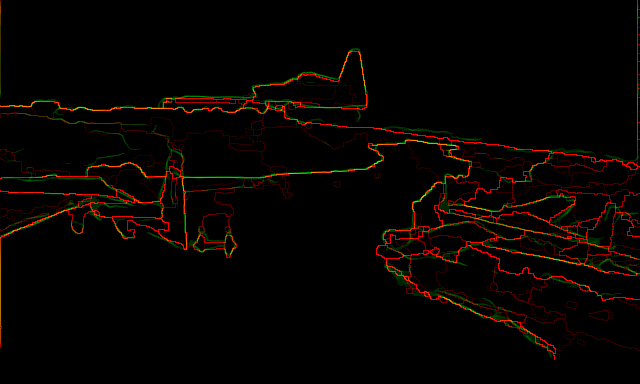}  
        \captionsetup{labelformat=empty,skip=0pt }
        \caption*{t + 4}
    \end{minipage}
    \centering
    \begin{minipage}{0.24\textwidth}
        \centering
        \includegraphics[width=1.0\textwidth]{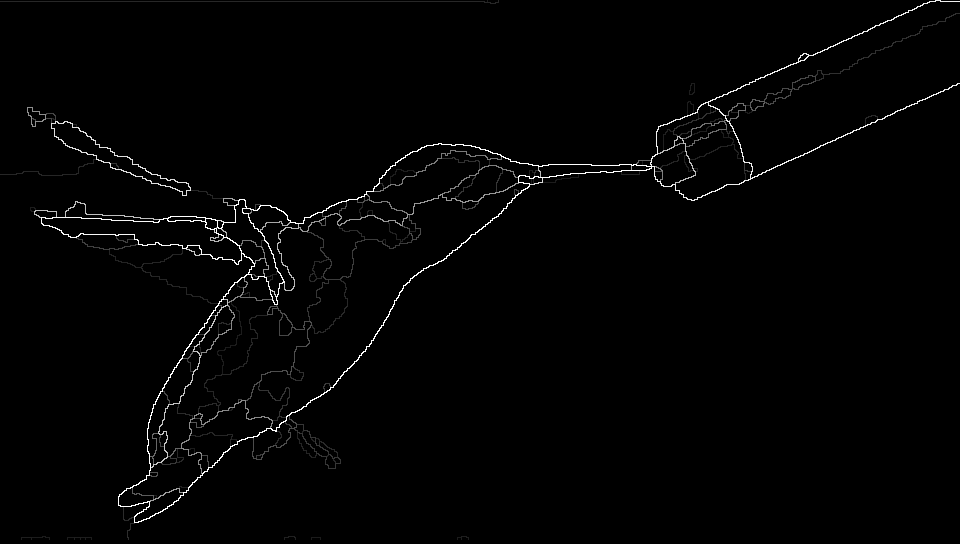}    
        \captionsetup{labelformat=empty,skip=0pt }
        \caption*{Last Input Frame}
    \end{minipage}
    \begin{minipage}{0.24\textwidth}
        \centering
        \includegraphics[width=1.0\textwidth]{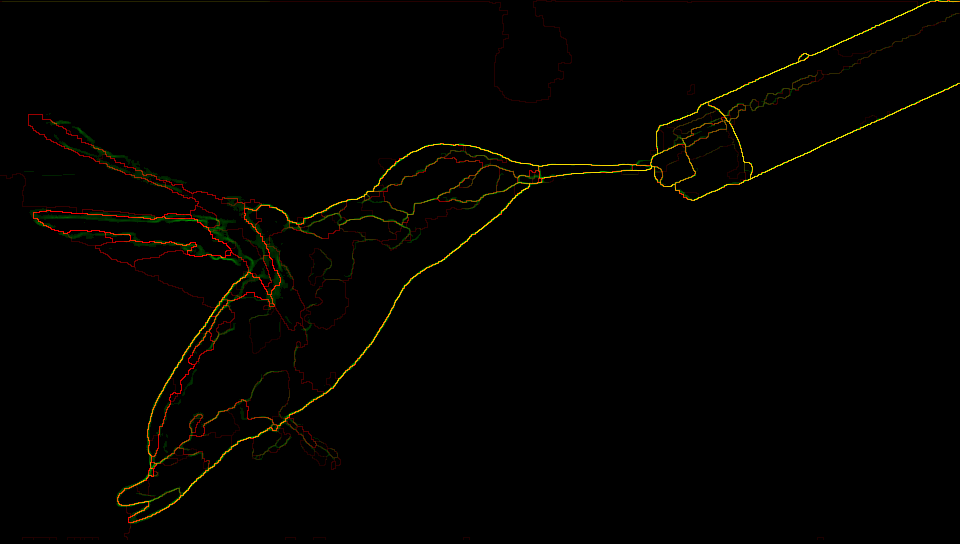}  
        \captionsetup{labelformat=empty,skip=0pt }
        \caption*{t + 1}
    \end{minipage}
    \begin{minipage}{0.24\textwidth}
        \centering
        \includegraphics[width=1.0\textwidth]{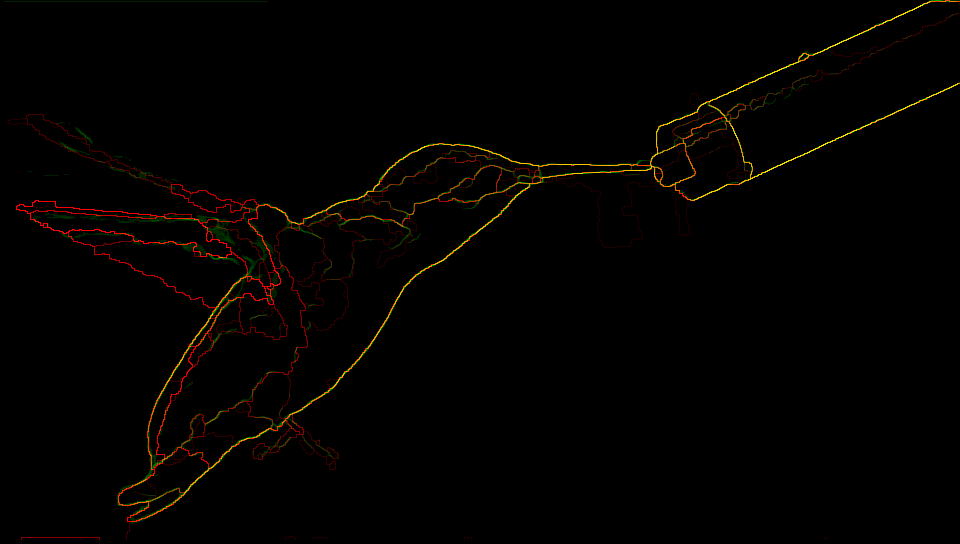}
        \captionsetup{labelformat=empty,skip=0pt }
        \caption*{t + 2}
    \end{minipage}
    \begin{minipage}{0.24\textwidth}
        \centering
        \includegraphics[width=1.0\textwidth]{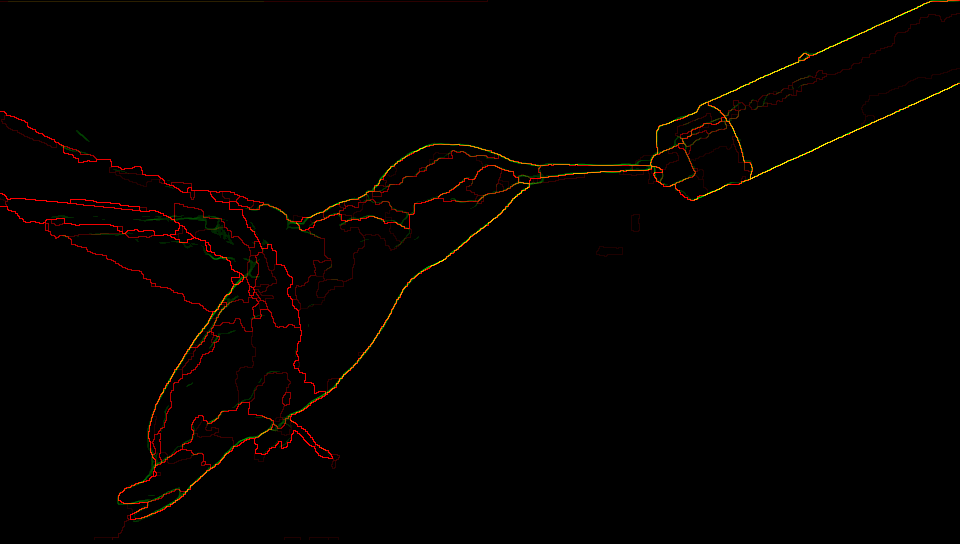}    
        \captionsetup{labelformat=empty,skip=0pt }
        \caption*{t + 4}
    \end{minipage}
    \caption{Columns top to bottom: Predictions on dominoes, airplane and hummingbird sequences from VSB100. In each predicted frame the actual boundary is coded in the red channel and extrapolated in green channel. Yellow means correct predictions. Extrapolated boundaries are usually sharp especially under smooth motion in dominoes and airplane. In the humming bird sequence, the models are not able to deal with the fast wing motion, but correctly extrapolates the slower moving legs.}
    \label{fig:frames}
\end{figure}

\paragraph{Discussion of results.}
We report the results in \autoref{fig:aucbestfvsb100} and \autoref{auc_fm_rec}. Overall, the Multi-Scale architecture (red lines) consistently outperforms the others. The Fully-Convolutional architecture (blue lines) is a close second. Adding convolutional layers (magenta lines) improves the performance of the RNN encoder-decoder architecture (cyan lines), however both have trouble beating the last input baseline. \\
\noindent
{\textit{Sequence to sequence vs Recursive (with and without context):} } As excepted the sequence to sequence prediction mode performs better than the recursive prediction mode especially at the t + 4 time-step. However, in both modes both precision and recall drop rapidly with time-steps.\\
{\textit{With or Without Context:} } Only the Multi-scale architecture is able to deal with high input data dimensionality induced by a context. The others, being simpler benefit from not having a context.\\
{\textit{Qualitative evaluation:} } Closer inspection of the results, suggests that the networks are not able to deal with large or unstructured motion. The networks in such situations react by blurring the boundaries, as a consequence of using the mean squared error. While predicting recursively this leads to loss of boundary confidence and eventual disappearance of the boundaries. However, the network is able to predict correctly in case of smooth motion e.g. the predictions in \autoref{fig:frames} from the videos airplane and dominoes.

\subsection{Synthetic data}
Dynamics of motion in the videos in the VSB100 dataset is frequently very complex and involve difficult to predict and non-deterministic actions of actors. To evaluate the performance of the models on structured, deterministic motion, we test the models trained on videos from the VSB100 on synthetic data. The synthetic data is sampled from worlds which consists balls moving on a frictionless surface with a boundary, akin to a billiard table. We used the pygame module of python to create such worlds and sample binary boundary images from them. As the target is always a binary image, we report only the best F-measure obtained by varying the output threshold parameter. Moreover, we consider only the recursive mode of prediction (with context) as the sequence to sequence mode and recursive mode without a context is limited by factors mentioned in \autoref{ssec:designchoices}.

\paragraph{Data generation from single ball worlds.}
\label{sssec:data_gen_syn}
We sample 30 sequences from the following parameters to create a synthetic test set:
\\
\textbf{Table size}: Side length randomly sampled from \{96,128,160,192,256\} pixels.\\
\textbf{Ball velocity}: Randomly sampled from [\{-3,..,3\},\{-3,..,3\}] pixels.\\
\textbf{Ball size}: Constant, with a radius of 13 pixels.\\
\textbf{Initial Position}: Uniformly over the table surface.

\begin{table}[h]
  \caption{Best F of models trained on VSB100 on single ball worlds}
  \label{auc_sb1_anna_30}
  \centering
  \begin{tabular}{llllll}
    \toprule
    \textbf{Time-steps} &\textbf{Last Input} &\textbf{RNN}   &\textbf{Conv-RNN}  &\textbf{Fully-Conv}  &\textbf{Multi-Scale}  \\
    \midrule
    t + 1      &0.141 &0.227 &0.397 &\textbf{0.612} &0.583  \\
    t + 2      &0.074 &0.119 &0.255 &\textbf{0.365} &0.352 \\
    t + 5      &0.038 &0.010 &0.017 &0.088 &\textbf{0.097} \\
    \bottomrule
  \end{tabular}
\end{table}

\paragraph{Evaluation of models trained on VSB100 on single ball worlds.}
We report the performance of the models in \autoref{auc_sb1_anna_30}. The models perform remarkably well on this data-set. The models are able to continue the motion of the ball even though they have never seen a moving ball before. However, they do not perform well near collisions and the predicted ball tends to slow down over time.

\subsection{Extrapolation over long time scales on billiard table worlds}
To further test the ability of the models to learn structured motion and extrapolate further into the future, we now train and evaluate the two best performing models from \autoref{fig:aucbestfvsb100} on billiard table worlds. 
\paragraph{Single ball worlds.}
 We generate a training set using parameters in \autoref{sssec:data_gen_syn}. However, to keep our training set as diverse as possible we prefer short sequences. We restrict each sequence to a maximum length of one or two collisions with walls and set a 50\% bias of the initial position of the balls being 40 pixels from the walls. We sample 500 such sequences and train our models on this from scratch. We then test the models on the test set generated in \autoref{sssec:data_gen_syn} and report the results in \autoref{auc_sb2_tr_ts}. We include as a baseline a ``blind'' Fully-Convolutional model, which cannot see the table borders. To beat this baseline, our models need to learn the physics of ball-wall collisions. We see accurate extrapolation even at 20 time-steps in the future.

\begin{figure}[t]
    \centering
    \begin{minipage}{0.24\textwidth}
        \centering
        \includegraphics[width=1.0\textwidth]{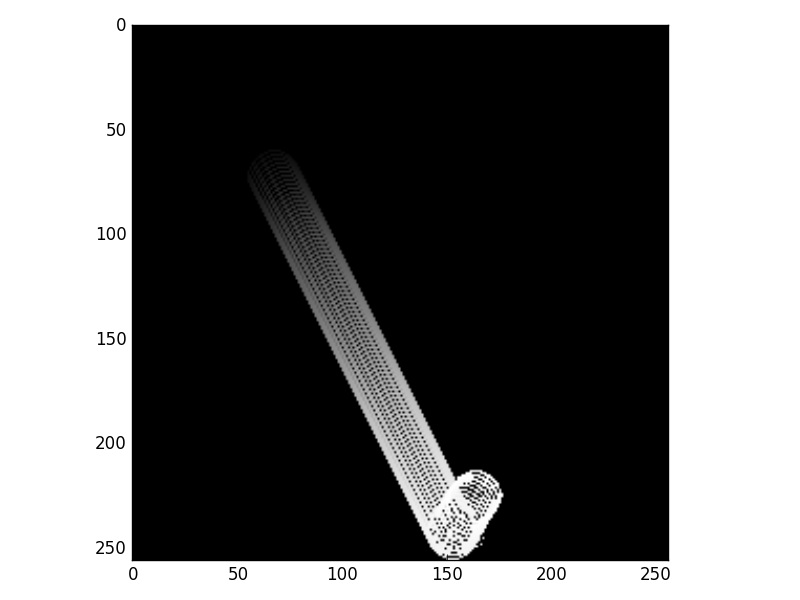}    
    \end{minipage}
    \begin{minipage}{0.24\textwidth}
        \centering
        \includegraphics[width=1.0\textwidth]{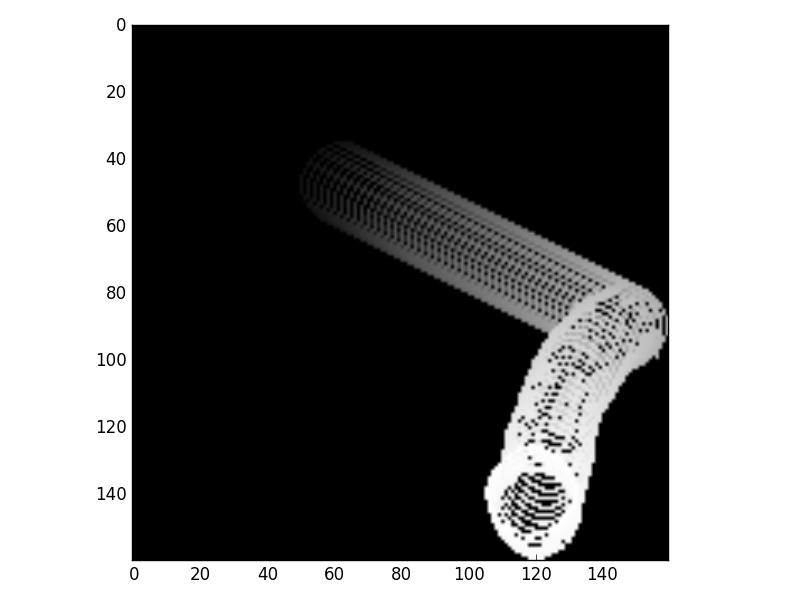}    
    \end{minipage}
    \begin{minipage}{0.24\textwidth}
        \centering
        \includegraphics[width=1.0\textwidth]{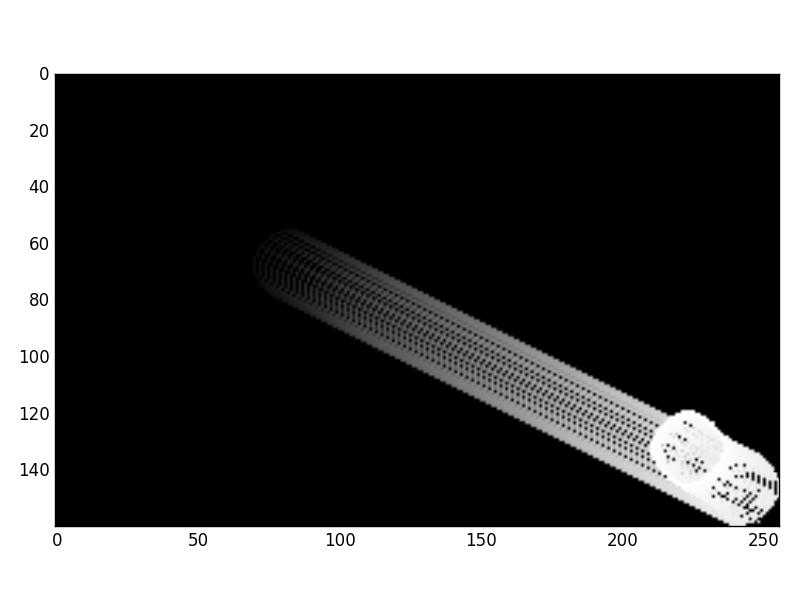}    
    \end{minipage}
    \begin{minipage}{0.24\textwidth}
        \centering
        \includegraphics[width=1.0\textwidth]{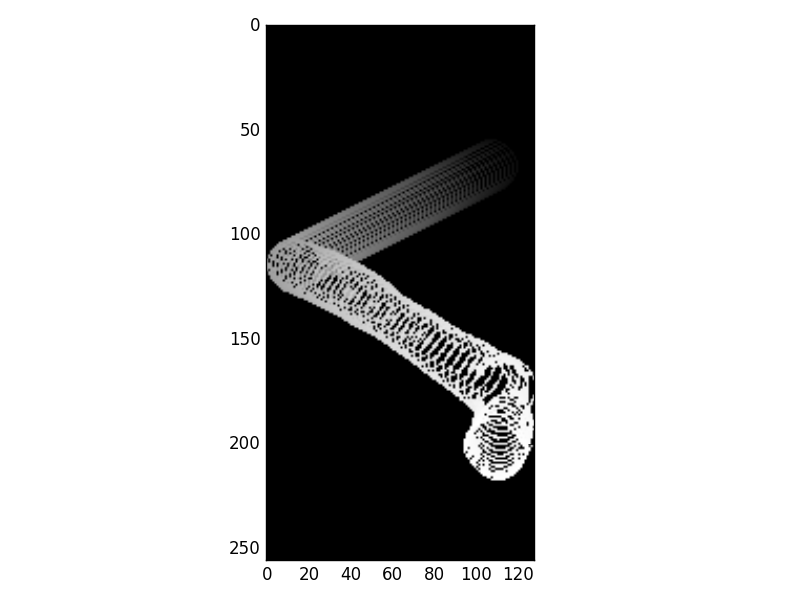}    
    \end{minipage}
    \begin{minipage}{0.24\textwidth}
        \centering
        \includegraphics[width=1.0\textwidth]{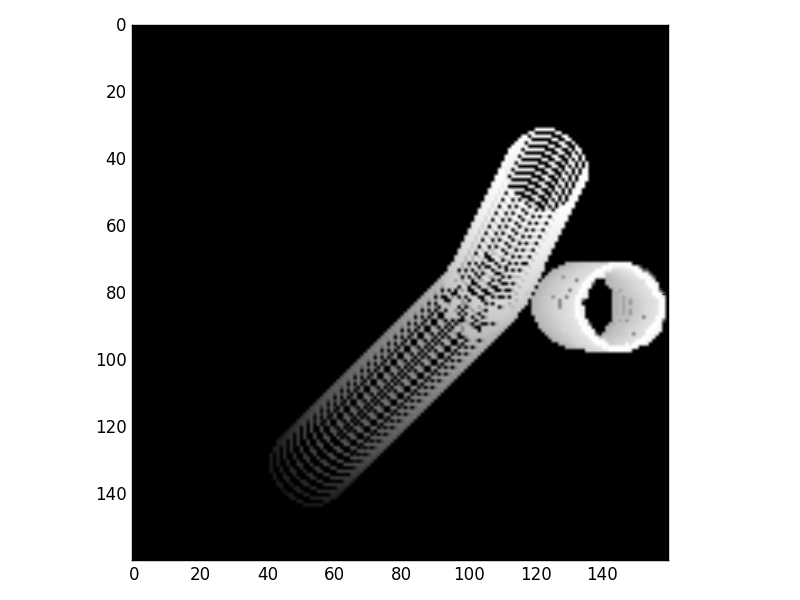}    
    \end{minipage}
    \begin{minipage}{0.24\textwidth}
        \centering
        \includegraphics[width=1.0\textwidth]{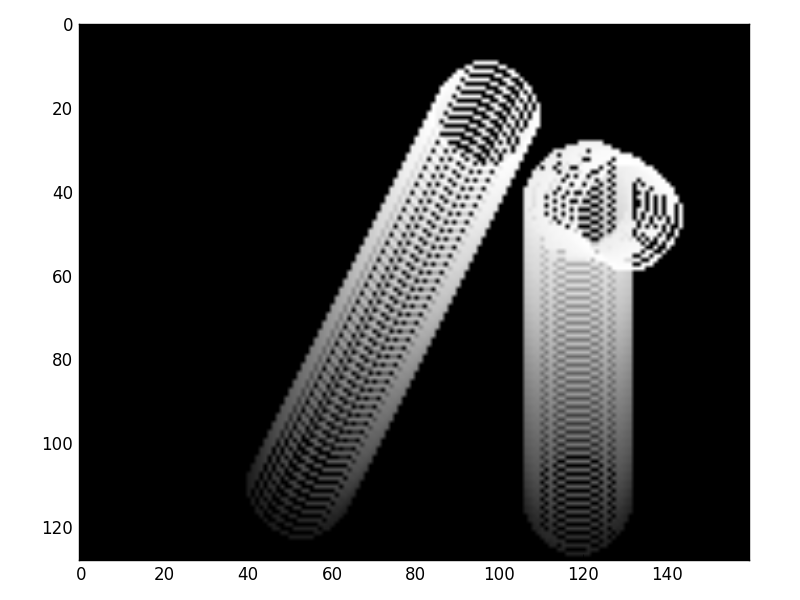}    
    \end{minipage}
    \begin{minipage}{0.24\textwidth}
        \centering
        \includegraphics[width=1.0\textwidth]{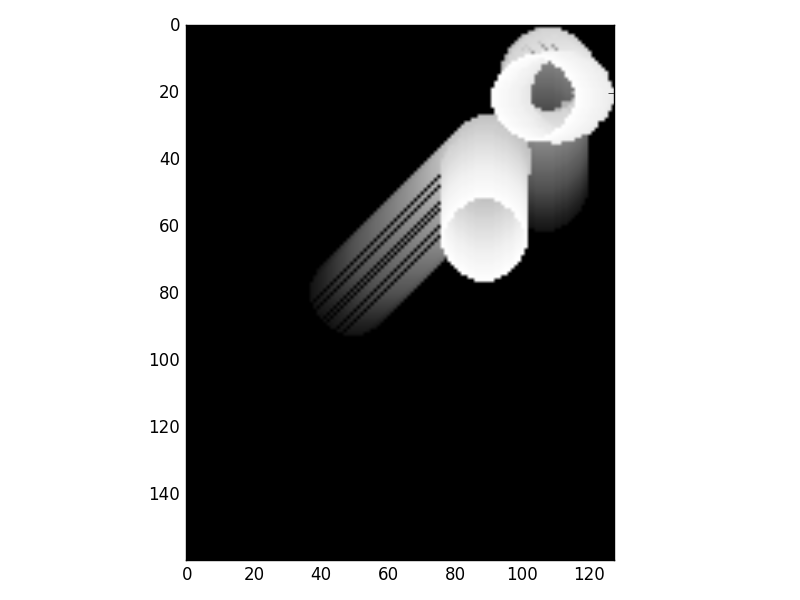}    
    \end{minipage}
    \begin{minipage}{0.24\textwidth}
        \centering
        \includegraphics[width=1.0\textwidth]{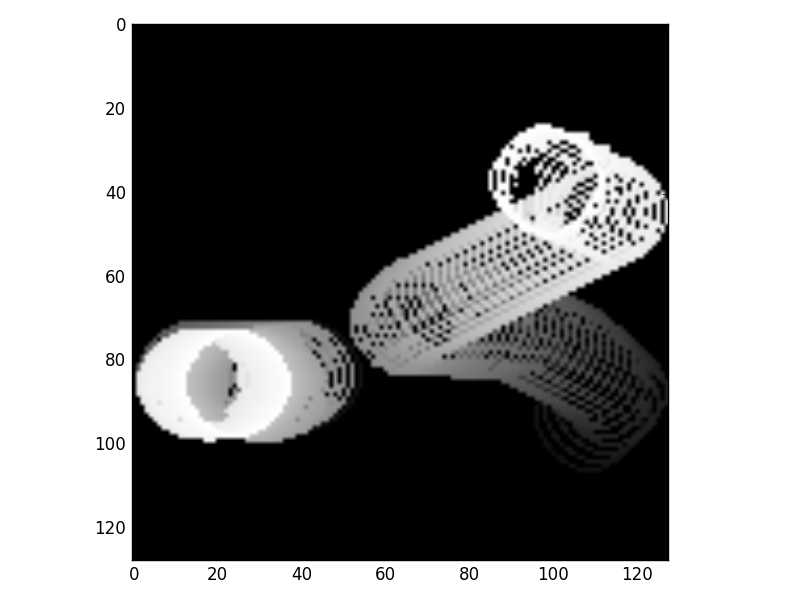}    
    \end{minipage}
    \caption{Trails produced by super-imposing extrapolated boundaries.}
    \label{fig:trails}
\end{figure}

\paragraph{Two and three ball worlds.}
Worlds with one ball only have collisions with walls. Worlds with more than one ball involve both ball-ball and ball-wall collisions, which make the physics of such worlds much more complex. To test the ability of the models to predict the future states of such worlds we sample training datasets which contains two and three balls respectively. We use the same parameters as \autoref{sssec:data_gen_syn}, except we limit the scene length to 200 frames. We sample 100 such sequences containing two balls and 50 containing three balls.  We use a curriculum learning approach, that is, we initialize the models with the weights learned on single and two ball worlds respectively.  We test the models on 30 sequences containing two and three balls respectively. We report the results in \autoref{auc_sb2_tr_ts}. We include as a baseline the Fully Convolutional models trained on single ball worlds and two ball worlds in the two and three ball world case respectively. This means the model has to learn the physics of ball-ball collisions beat the baselines. This also allows us to evaluate models trained on a simpler world on a more complex one. Again, we see accurate extrapolation even at 20 time-steps in the future.

\begin{table}[h]
  \caption{Evaluation of models trained on synthetic data (training on the same world)}
  \label{auc_sb2_tr_ts}
  \centering
  \begin{tabular}{llllll}
    \toprule
    \textbf{Time-steps}  &\textbf{Last Input} &\textbf{Baseline} &\textbf{Fully-Conv}  &\textbf{Multi-Scale}  \\
    \midrule
    \multicolumn{5}{c}{Best F: Evaluation on single ball worlds}                 \\
    \midrule
    t + 1      &0.141 &0.964 &\textbf{0.994} &0.987 \\
    t + 5      &0.038 &0.791 &\textbf{0.956} &0.900 \\
    t + 20     &0.002 &0.637 &\textbf{0.709} &0.632 \\
    \midrule
    \multicolumn{5}{c}{Best F: Evaluation on two ball worlds }                 \\
    \midrule
    t + 1      &0.246 &0.941 &0.951 &\textbf{0.969} \\
    t + 5      &0.114 &0.776 &0.848 &\textbf{0.896} \\
    t + 20     &0.101 &0.545 &0.566 &\textbf{0.681} \\
    \midrule
    \multicolumn{5}{c}{Best F: Evaluation on three ball worlds }                 \\
    \midrule
    t + 1      &0.246  &0.950 &\textbf{0.969} &\textbf{0.968} \\
    t + 5      &0.118  &0.823 &0.864 &\textbf{0.892} \\
    t + 20     &0.090  &0.550 &0.585 &\textbf{0.700} \\
    \bottomrule
  \end{tabular}
\end{table}

\paragraph{Extrapolation over very long time scales.}
Although we evaluate only 20 timesteps into the future in \autoref{auc_sb2_tr_ts}, our models are stable over longer time-horizons. We gave the first 4 frames as input and asked the models to extrapolate 100 frames into the future. We superimpose the frames to produce the trails in \autoref{fig:trails}. However, we noticed that sometimes the balls reverse direction mid table and the ball(s) get deformed or disappear altogether. 

\section{Conclusion}
We propose a new challenge of predicting boundaries in future video frames as well as several architectures for this novel problem of boundary extrapolation. We investigated a range of design choices and  found that our ``Multi-Scale'' architecture works best on both real and synthetic data. In contrast to prior work, we observe that our model formulation obtains accurate and sharp results even with mean squared error. This lends support to our claim that boundary extrapolation is indeed a better behaved problem than natural frame prediction. Moreover, accurate results on varied scenarios involving billiard balls shows that our models can develop an intuitive notion of physics as well as could lend itself to formulating expectations over future frames in advanced video segmentation methods.

\newpage
\begin{spacing}{0.9}
\small
\bibliography{refs}
\end{spacing}

\end{document}